# Beyond the Org Chart: AI and the Transformation of Invisible Work

**Stephanie Rosenthal and Shamsi Iqbal**
**Microsoft Corporation**
**{srosenthal, shamsi}@microsoft.com**

**Abstract**

An increasing number of news and research articles report that AI adoption is allowing professionals to blur and extend the boundaries of their corporate roles. With the goal of understanding how work processes might be changing in an AI-forward company, we interviewed 24 product-focused individuals at a large technology firm about how AI has impacted their own work, their work within their product team, and their professional interactions. Our conversations suggest that AI is not only changing formal role responsibilities and collaborations between those roles, but also changing informal cultural practices like professional mentoring that are key to helping professionals settle in their positions, stay engaged with their work, and grow their careers. Some of these changes are positive, such as smoother collaboration between peers, but other changes are more nuanced and put the typical career growth opportunities, like receiving feedback from professional networks and promoting leadership and mentorship, at risk. We propose steps that AI companies can take to make the invisible work more visible. Additionally, we propose efforts that individuals and leaders can take to support their colleagues through AI transformation while preserving healthy company cultures that support diverse thinking, collaboration, and informal interactions.

## Introduction

The rapid integration of artificial intelligence (AI) into knowledge work industries is fundamentally transforming professional productivity, task structure, and decision-making processes. Empirical research shows that generative AI can significantly enhance worker performance across a variety of tasks. For example, one recent study showed that agents could complete tasks ~90% faster than humans on the same set of 16 tasks (Wang et al. 2025). Another large-scale study of over 5000 customer support specialists found that AI-assisted workers completed more tasks, delivered higher-quality outputs, and worked faster than those without AI assistance (Brynjolfsson, Li, and Raymond 2025). Despite the disagreements in exactly how much productivity improvement workers see, both workers (Woodruff et al. 2024) and the media (Vallance 2023) foresee the monumental impact of AI on "visible" work that directly affects an organization's bottom line.

Emerging research indicates that AI may also affect individual workers' cognitive, psychological, and job roles (Oliveira, Carvalho, and Faria 2025; Wolfe, Choe, and Kidd 2025; Chang et al. 2024). Job tasking is expanding and changing as AI enables new skills, and these changes are, in turn, affecting workers' anxiety over roles as well as their sense of empathy for their coworkers (Oliveira, Carvalho, and Faria 2025). AI automation is impacting role changes on workers' job satisfaction (Ranjit et al. 2026). Additionally, several studies have found that AI may induce technostress symptoms among workers, and have tracked coping mechanisms that they are adopting to avoid burnout with the increased intensity of work (Chang et al. 2024; Kwon et al. 2026).

Despite the growing body of knowledge surrounding AI's impact on visible work outcomes and personal well-being, there is less known about the effects on the relationship-building, support-giving, and work culture-boosting *invisible* labor (Daniels 1987; Hatton 2017) mechanisms that are socially and structurally necessary for healthy organizations (Zabbo Feb 3 2026). Examples of invisible work include conventional social support (House and Kahn 1985) such as providing help, information, and feedback, but also the efforts taken to maintain an open and welcoming community including onboarding new workers and developing common ground necessary for smooth collaborations.

In order to understand the impact of AI on invisible work, we performed semi-structured interviews in March and April 2026 with 24 professionals working across different job roles at a large AI-first technology company that has been emphasizing the use of AI across a broad swath of job tasks. In one hour discussions, we asked each participant to describe the workflows they perform on a regular basis and where they used AI, how they used to perform their role without AI, and how their job roles have changed over the last 6 months. We then performed an analysis of the transcripts to find common themes around AI impact on personal work output and collaborative or group projects, especially those interactions and tasks that are not defined in their formal roles.

We found that in addition to validating formal role changes that have been observed in other research (Oliveira, Carvalho, and Faria 2025), the invisible work associated with the roles is also changing across seven dimensions. AI is enabling new broader skill sets for professionals transcending their immediate roles, which, in turn, are (1) changing the language of communication between different roles and also (2) blurring the boundaries between job roles and creating anxiety. It is (3) allowing workers to be more efficient about creating situational awareness around projects, and (4) requiring that they manage their reputation and personal accountability for the work they are creating. Finally, workers are using AI tools to (5) support their own learning and growth as well as (6) get help and feedback for their tasking and (7) receive proxy peer collaboration support, but at the risk of losing the professional networks that are critical for career advancement. Overall, our participants feel more independent and empowered to do their work using AI but are also missing out on human social support, putting them at risk for work loneliness, slowed career growth, burnout, and attrition (Hadley and Wright 2024, 2026).

Based on these findings, we conclude that, in the workplace, AI is impacting the invisible work just as much or more than the visible work. We discuss opportunities for AI systems to encourage interactions among collaborators to balance workers' increasing adoption of AI tools that may socially isolate them, potentially helping them build and maintain critical professional and support networks. We also discuss opportunities for both individuals and leaders to support their colleagues by offering help and feedback, opportunities to connect with peers, and holding them accountable for their AI-augmented work.

## Related Work

Original research on invisible work focused on domestic work traditionally managed by women and mothers (Daniels 1987), but has been extended in recent years to a variety of other tasks that typically also fall on women and minorities. In the workplace, invisible work is the term used to describe tasks that are necessary for well-functioning groups or organizations, but often go unnoticed, untracked, and especially unrewarded (Hatton 2017). The types of tasks that fall into invisible work can roughly be categorized as interpersonal, social, and cultural in nature, and include emotional support, conflict resolution, mentoring, transfer of knowledge, office organization, communication, coordination, building rapport, behind-the-scenes networking to build and maintain professional relationships, and coaching others (Wells and MacAulay; Time 2023; Szulanski 1996). Invisible work has real impacts on workers' stress, job satisfaction, job performance, happiness and self-worth, and leads to blurred work-life boundaries and strained personal relationships (Daniels 1987; Time 2023) as well as organizational information flow, learning and solving complex tasks (Szulanski 1996; Brown and Duguid 1991; Moreland 1996).

Many of the invisible work activities are centered around social support. Research on work and well-being has long-established social support as a core resource rather than a peripheral benefit in organizations. Early stress research defines social support as resources provided by others that help individuals cope with demands, particularly by shaping how stressors are appraised and managed (Cohen and Wills 1985). In organizational contexts, support consistently predicts lower strain, higher satisfaction, and better performance, especially under high demands (House 1983; Viswesvaran, Sanchez, and Fisher 1999). Contemporary work-design theories formalize this insight: both the Job Demands–Resources (JD-R) model and related frameworks treat social support as a key job resource that buffers burnout and sustains engagement when workloads and uncertainty increase (Demerouti et al. 2001; Bakker and Demerouti 2017).

The literature converges on a small set of support types that structure workplace relationships. Classic typologies distinguish task support (tangible help and task assistance), informational support (advice, guidance, and knowledge), emotional support (empathy, reassurance, care), and appraisal or feedback support (affirmation, evaluation, and coaching) (House 1983; Cohen and Wills 1985). A meta-analysis shows that these forms are related to a variety of outcomes: instrumental and informational support are strongly tied to performance, while emotional and appraisal support are more closely linked to well-being, commitment, and reduced withdrawal (Viswesvaran, Sanchez, and Fisher 1999; Ng and Sorensen 2008). Importantly, many benefits of support derive not only from problem-solving but from relational signals of belonging and legitimacy, which shape identity and motivation at work (Kahn 1990).

Against this backdrop, recent practitioner and academic discussions highlight a shift: workers are increasingly turning to generative AI (e.g., large language models and conversational agents) for forms of support traditionally provided by coworkers or managers (Hadley and Wright 2026). Recent empirical work shows that workers use generative AI for informational and instrumental support such as drafting, debugging, sense-making, and learning, often with substantial productivity gains, particularly for less experienced workers (Brynjolfsson, Li, and Raymond 2025). Other studies similarly find that developers and knowledge workers value generative AI as an always-available source of guidance and task scaffolding, while drawing boundaries around relationship-centered activities such as mentoring and evaluation (Barke, James, and Polikarpova 2023; Choudhuri et al. 2025). In a review of AI use in human-resource management, further analysis finds workers turning to AI for everything from knowledge sharing to cooperation to collaboration with effects on job satisfaction and job placement (Pereira et al. 2023).

Workers also use generative AI for emotional and appraisal-like support, including reassurance, confidence building, and career advice. Research on conversational agents shows that people readily attribute empathy and understanding to systems that produce fluent, responsive language, even when they know those systems are artificial (Nass and Moon 2000; Pelau, Dabija, and Ene 2021; Zhang et al. 2025b). This helps explain why AI-mediated support can feel psychologically meaningful in the moment. However, evidence suggests a tension between perceived support and social outcomes: while AI can simulate responsiveness, it does not reliably reduce loneliness or replace the relational benefits of human connection (Campbell 2021). From a classic support perspective, this implies that AI may deliver informational content without fully reproducing the stress-buffering and identity-affirming functions of human support (Hadley and Wright 2024, 2026).

Recent organizational research further complicates substitution narratives with a series of paradoxes outlining both the benefits and costs of AI at work. Studies of AI use at work show that workers may incur a social evaluation penalty when coworkers learn they rely on AI, being judged as less competent or less trustworthy, especially when AI use is disclosed without clear norms (Zhou et al. 2025; Reif, Larrick, and Soll 2025). This creates a paradox: generative AI can increase individual efficiency while undermining relational standing, which is itself a pathway to support. Similarly, dependence on AI at work may simultaneously augment human capabilities and also reduce human expertise (Ehsan et al. 2026).

Taken together, the literature suggests that generative AI is a double-edged sword. It can increase productivity on individual tasks, but also may induce stress through the increased intensity of work (Ranganathan and Ye 2026). In regards to social support, workers are increasingly using AI to scale instrumental and informational support and provide low-stakes collaborative scaffolding, it is unclear how it affects interpersonal relationships that are critical to career success and organizational effectiveness. The emerging challenge for organizations is therefore not whether workers will use AI for support, but how to measure the impact of that support on the organizational structures that have typically gone undocumented and how to design norms and systems on top of them that preserve human connection while leveraging AI's accessibility and reach. In this work, we investigate the impacts of AI on invisible work within a large technology company, outlining the benefits and risks across seven key impact areas, and propose steps that AI companies, individuals, and organization leaders can take to support their workers and maintain a healthy work culture.

## Research Methods

In order to understand the impact of AI adoption on invisible work, we performed semi-structured interviews with 24 professionals at a large technology company. We chose to investigate AI impacts on work at a company that, like many others (Nahar et al. 2025; Choudhuri et al. 2025; Butler et al. 2025), has provided its professionals many different AI tools for an extended period of time. Our expectation is that long-term adoption has allowed our participants to find consistent patterns of use that are successful for them and their product teams, and they will be able to reflect on their sustained use and identify more subtle work changes.

The interview participants were asked to describe common workflows they accomplish regularly, and how those workflows have (or have not) changed due to AI. We analyzed the transcripts for common themes related to visible work (e.g., productivity and deliverables) as well as to invisible work (e.g., collaboration, help, and feedback). Our results show that AI adoption has both positively and negatively impacted invisible work, which in turn has the potential to impact social support, individual career growth and company culture in the future.

| Profession | Total | Male | Female |
|---|---|---|---|
| Design | 7 | 4 | 3 |
| User Research | 7 | 4 | 3 |
| Applied/Data Science | 10 | 5 | 5 |

Table 1: Count of participants from recruited professions.

### Participants

Participants were recruited through emails sent to managers of sociotechnical professions - designers, user experience researchers, applied scientists, and data scientists. Unlike software engineers whose AI tooling has been largely focused on coding tasks, participants in these professions have access to a large suite of different tools that they must sift through and choose whether to adopt for their particular product and tasking needs.

In order to participate in our IRB-approved study, participants must have been hired at the company at least 6 months prior to the study and have the job titles listed above. The recruitment emails gave recipients the opportunity to take a survey and also sign up to be contacted for our deeper interview, the results of which are presented in this paper. After reading the consent form for both studies, the participants could enter their email address into the form if they agreed to be contacted about the 1-hour interview. Regardless of whether they entered an email address, they were given a link to the survey.

In total, 24 participants agreed to be interviewed during the months of March and April 2026. Table 1 outlines the counts of participants across each profession and gender. Three of the participants were managers and the rest were independent contributors. Four participants (two design, one user researcher, and one applied science) self-reported as avid AI users, and one was particularly pessimistic about AI use in their job. To maintain the privacy of participants within the organization, no further demographic information was collected.

### Data Collection

The interviewer started every interview with an outline of the questions that would be asked in order to prime participants for what to talk about. The participants were also told that we were more interested in the work they were doing and not the specific tool that helped them achieve the work. This helped limit the amount of feedback we received about which AI tools were better than others, and instead focused on the benefits and challenges of tools in general. Interviews took place over video chat and, just before starting the questioning, participants were given the option to allow us to use built-in transcription services or opt out of transcription[1]. Twenty-two participants agreed to our use of AI transcription during the interview.

Participants were first asked to describe several workflows that they perform fairly often. The workflows chosen varied greatly by participant and included areas such as:

- writing and content generation,

[1]In this case, the interviewer took notes manually.

- literature reviews and product/company research,
- visual and interaction design,
- coding and vibe coding,
- interviews, surveys, and other research methodologies,
- evaluation and data analysis (AI and human-generated).

The interviewer asked probing and clarifying questions about what aspects of the tasks used AI, if any. After hearing about each workflow, the interviewer asked the participant to reflect on how that workflow was done before AI was introduced and how accurate the AI tools are in helping them complete their tasks. With respect to non-AI methods, participants frequently discussed lengthy interactions and iterations with peers and more time spent thinking before execution. At times, they shared workflows that were new skills acquired with AI and were unable to complete the task without it i.e., coding in a particular language or using some other skill set that they never learned.

Next, reflecting on AI use in general, participants were asked to discuss first how AI affected themselves or their work either positively or negatively. Here, participants commonly mentioned being more productive, spending less time iterating with peers, feeling more independent, feeling more or less creative, and feeling more socially isolated when they cannot find the help that they need. Then, they were asked to describe how AI affects their product team or their professional discipline team. Common answers included reflections on collaboration, on learning and sharing about AI together, and thoughts about AI integration into products.

Finally, the interviewer asked participants to reflect on how their job roles and tasking changed in the last 6 months, whether and how they expect continued changes in the next 6 months, and whether AI has affected their sense of job security. Participants frequently talked about how their job tasks have expanded to a broader set of objectives or how the intensity of their work has changed. Some participants, especially the designers and user researchers, felt that their roles had changed a great deal, while the data and applied scientists did not see much shift in their roles' goals. For job security, all participants felt that jobs in the technology sector were never secure, but no one felt their position or their job role was in imminent danger. The general consensus was that they did not know what the future of work would be like in 5-10 years, and that positions may be completely different by that time.

## Transcript Analysis

After completing each interview, the interviewer collected the transcript, cleaned and anonymized it, and tagged it based on each question asked and any other relevant themes that emerged from the conversation. Due to the private nature of the discussions which at times included brief references to coworkers, leaders, or customers, references to particular products, teams, and people were removed and replaced by tags like [Company X], [Company Y] (if distinct), [Participant A] (self), [Manager] or [Colleague], and [Product]. Themes outside of the question titles that were tagged included independence, creativity, impact, isolation, and collaboration.

We performed a secondary review of each transcript after more than 80% of the interviews were complete. We tagged important quotations that described the participant's sentiment on a particular topic, and common themes that were repeated among multiple participants, but not necessarily called out in the initial review. The themes around institutional knowledge and help/feedback theme emerged later in our review. Finally, we searched for correlations between our themes. For example, we noted that some participants, particularly the designers, mentioned collaboration and communication a lot while others discussed their interactions with other teammates very little. In conjunction with the observations around reputation, accountability, and feedback, these findings were significant and offer insights into the diverse perspectives that professionals have about AI and the impacts on their ability to get work done.

## AI Transformation of Invisible Work

There is a tremendous amount of informal invisible work that our participants perform to manage their relationships with their peers and leaders, including:

- communicating effectively across disciplines
- sharing project status and outcomes with colleagues
- informally leading and mentoring co-workers
- building rapport and collaborations with peers
- providing help and feedback on projects

While some of these tasks focus on productive collaboration, others involve social support including providing and receiving help and feedback. Maintaining effective relationships with colleagues is crucial for building professional networks, mentoring, and career growth opportunities. The lack of social support at work has been linked to feelings of loneliness, and lower job satisfaction and productivity (Hadley and Wright 2024, 2026). Our thematic analysis uncovered seven ways that AI has impacted invisible work for our participants. The themes are summarized in Table 2 and counts of the number of participants who discussed those themes are shown in Table 3.

### 1. Cross-Discipline Communication

*is smoother as AI supports the creation of new handoff boundary objects, and AI chats help explain complex concepts.*

Diverse opinions are key to building successful products (Post et al. 2009), yet sometimes collaboration among colleagues with very different backgrounds makes it challenging for them to find common ground (Zhang et al. 2025a). Our participants reported that reaching consensus about interaction approaches has been a point of friction within some product groups. This is because there can be a disconnect between what designers envision as the best possible customer experience and what engineers feel they can feasibly accomplish given the legacy code bases they work with and the time allotted for making new features.

Designers said they were particularly impacted by AI because they are increasingly able to create high-fidelity prototypes more easily which allows them to establish common ground with developers. As new AI coding tools have allowed for 'vibe coding' using natural language, designers broaden their skill set to include front end user interface design and engineering (Hohn and Loydl 2026). Participant H remarked that they feel a sense of empowerment in being able to focus implementing prototypes rather than visual experiences, and that vibe coding has allowed them to create rich proofs of concept for engineers:

> There's always been a little bit of friction between design and development. I used to come with these crazy visions and they're like, "yeah, sure, buddy, you know, but we can never actually implement any of it." And now that I can show them, "look, it's actually code."

| Invisible Work | Impact |
|---|---|
| Cross-Discipline Communication | New AI-driven boundary objects are smoothing interactions with peers |
| Role Clarity & Division of Labor | The blurred boundaries between disciplines are empowering and anxiety-inducing |
| Project Situational Awareness | Overhead to revise and share information and artifacts is cumbersome |
| Reputation and Accountability | Readers do not hold writers accountable for their written content |
| Personal Growth and Learning | People are learning new skills through interactions with AI |
| Instructional Support and Feedback | AI tools provide fast access to information at the cost of human support |
| Task Support and Collaboration | Increasing pace of work is affecting collegial support and opportunities for AI peers |

Table 2: Our analysis identified seven invisible work areas that are impacted by AI.

Prototypes act as boundary objects that have been shown to smooth collaboration between disciplines (Gunasekaran et al. 2026), and our participants have discovered that they can collaborate more closely with software engineers through shared code (Meske et al. 2025; Zhang et al. 2026).

The high-fidelity prototypes are also changing communication with the product designers (PMs) and even customers. Participant J showed off a new system for version-controlling the prototypes as they are updated.

> And this is what we're sending to the PMs and the devs and being like, "hey, you can look at here. We're updating this probably everyday now, just check, take a look." You can show these to the customers too.

The tools are changing the bottlenecks in the development pipeline, enabling designers to show prototype demos live so that developers can interact with them to clarify desired interaction behaviors and PMs and user researchers can show them to customers to get feedback earlier.

Changing handoff boundary objects extends beyond designers to other fields as well. Participant X, a user researcher, talked about creating markdown files instead of presentations to share findings:

> I asked the PM what would be helpful for you to work, and have an effective working relationship with researchers? And she said make your research more digestible. And so, this morning, I was turning our heuristic guidelines into markdown files.

The markdown files are more easily searched and included as context for AI to use for subsequent steps in the development process. Additionally, Participant G explained how AI assistants can also be used to help explain difficult engineering concepts without the designers needing to explicitly ask for clarification.

> [AI is] shortening the gap between engineering and design, because designers now can just ask the agent what engineers are talking about

By utilizing AI for technical explanations, designers do not have to risk embarrassment by asking their colleagues and can participate more fully in collaborative decision-making meetings.

## 2. Role Clarity and Division of Labor

*is needed as AI enables new adjacent skills and blurs the boundaries between disciplines.*

In collaborative cultures, role clarity has been linked to higher job satisfaction and productivity as well as lower anxiety at work (Gil-Garcia et al. 2019). AI transformation is affecting role clarity as workers broaden their work scopes by acquiring or benefiting from new AI skills (Wolfe, Choe, and Kidd 2025). Recent findings suggest that one determining factor in whether workers understand the division of labor between disciplines or groups is clear boundary objects, which, as discussed above, are currently in flux in the technology industry, further exacerbating the anxiety that people are feeling (Gil-Garcia et al. 2019).

New AI skill sets have allowed our participants to envision new job roles in which design and front-end development are performed by a single person (Zhang et al. 2026). Because the legacy code bases are not built on modern technology stacks that the AI tools are best capable of producing, we also heard frustration from designers over their continued lack of independence with respect to being able to contribute to the code bases on their own. This frustration seems to be in conflict with the sense of pride and empowerment in the new-found AI vibe-coding skills. Participant I said:

> [Submitting code] changes by themselves through our products becomes a very complicated process ... by the time they get to fix that bug, it requires so much investigation and digging because the code wasn't written in the modern stack, or it wasn't the clear separation of front end and back end.

| Invisible Work | Count |
|---|---|
| Cross-Discipline Communication | 14 |
| Role Clarity & Division of Labor | 14 |
| Project Situational Awareness | 14 |
| Reputation and Accountability | 13 |
| Personal Growth and Learning | 11 |
| Instructional Support and Feedback | 11 |
| Task Support and Collaboration | 10 |

Table 3: The count of participants who spoke to each theme.

Sustained efforts by our designers to submit code to the software repositories has led to the accurate realization that software development at a large company is complex, and the boundary objects that AI are helping them produce are not without friction. This friction may stem from the disconnect between the publicity around "AI-first" development practices being fast and accurate, and the purported shortcomings of vibe-coding including the lack of scalability and testing that are necessary for long-term deployment of software systems to diverse customers (Fawzy, Tahir, and Blincoe 2025). Software engineers still play a significant role in developing software systems, even though they are augmented by AI agents (Abbas et al. 2025; Butler et al. 2025). AI today does not make up for the collection of skills beyond code writing that software engineers perform.

Taken at a higher level, these findings also indicate lack of clarity in job roles and division of labor as AI allows different professions to expand their skills. Many of our participants asked themselves what makes particular job roles unique and important, and what will happen if some roles are deemed irrelevant or unnecessary. Participant C spoke about the fact that both designers and developers are asking the same questions:

> When we got designers programming, so they're like, is our role kind of not around anymore if we're going to be developers? And engineers are going, well, if they're coding it, then do they need us?

Organization leaders have the opportunity to help drive role clarity and reduce this anxiety that is affecting collaborative culture.

### 3. Project Situational Awareness

*is simultaneously easier to create through AI writing tools, but also harder to operationalize as fewer people are reading more documents.*

Building situational awareness, rather than withholding information, is an important part of working in collaborative and open environments (Webster et al. 2008). Situational awareness is key to high-functioning team performance, as teammates create shared mental models that allow them to move quickly and with fewer misunderstandings (Salas et al. 1995). Significant time is spent preparing documents and meeting to share information (Baksa and Branyiczki 2023). Sharing information with leaders and managers is important to maintaining visibility and impact of the work presented (Zhang, De Pablos, and Zhou 2013), which affects the planning of future project work and the professional growth of the writer. Additionally, sharing outcomes on wide distribution lists provides "lurkers" the opportunity to hear about information they would not have otherwise been privy to (Cranefield, Yoong, and Huff 2015), expanding their working knowledge for their own projects, their professional networks, and potentially their skill sets.

Participants in our study said that their teams spend a lot of time sharing information about their project status and project outcomes both internally and across broader organizational structures. Sharing context can take many forms including meetings, project presentation share-outs, written documents, and status emails. Participant U spoke about the work to build situational awareness continuing in similar ways to before AI:

> I think that interpersonal "getting work done" with others is still uniquely human and it just is clunky. I mean, it's just going to be clunky in some ways. Like you still have to navigate time zones. You still are putting things in documents and people are making comments on them... I think we are all still human and we still work similarly.

They continued to say that the work to build a shared representation of knowledge still takes time, and negotiation remains a bottleneck in collaborative work. While the challenges of communication have not changed, Participant J commented on how the pace of work has shifted due to AI and the awareness of work that their teammates are doing has also been impacted:

> Now I have to ask what everyone's doing in their own AI exploration. I'm starting to be a little worried that we're moving so fast that we're not looking side to side.

They spoke about how they need to be more deliberate in asking their colleagues ("side to side") what they are working on to build situational awareness.

With regards to AI tools supporting situational awareness, one nearly-universal finding across all participants is that writing the documentation necessary for sharing knowledge is easier with AI. Participant F said "I would say it's empowering.... Like before [AI], I don't think I could write extensive articles" and they would have needed extensive support from colleagues or supervisors. Now, they write long research articles with the help of AI.

Participant V feels like their audience is not reading the content they generate any more:

> It's just really hard to work with the attention spans that were already limited before AI, but now it's like shrunk where before people would be okay reading a one-pager document. Now they want like 2 bullet points, 3 bullet points. But it's like "how do I make sure you're informed enough to actually incorporate those two bullet points correctly?"

They went on to talk about instances where their collaborators did not take their research into account or where they had to constantly repeat their points because peers were not paying attention. They also described strategies for distilling information in novel ways including short video clips, presentations, and markdown files. Participant N made an agent that their colleagues could ask about prior research results rather than having to spend the time re-sharing.

Some participants felt that they were using AI to produce documents that others would use AI to summarize, creating an artificial pipeline of information that risks missing the true meaning of the work. Recently, researchers have also argued that the emergence of "AI slop" is an artifact of the increase in the supply of written documentation, and despite the frustration, there is positive value in the content and will affect the "creative, information, and cultural" economies of consumption (Kommers et al. 2026). However, based on our study, it seems that workers are adapting their writing styles to more concise modalities, which may have a much broader impact on productivity and project deliverables if they cannot mitigate the risks of misunderstandings.

## 4. Reputation and Accountability

*are key drivers of ensuring AI content quality, and individuals must hold themselves and their peers accountable by calling out and questioning what they read.*

Building a high reputation at work is another invisible work task that takes time and responsibility to maintain, and also impacts the way that work tasks are completed. Research shows that people who have higher reputations feel a stronger sense of accountability and less tension and strain holding themselves to that standard than those with lower reputations in organizations (Laird et al. 2009). While accountability is often placed externally (i.e., others hold a person accountable for their actions), there is also recent work showing that people who have internally-placed accountability are more likely to take on additional organizational tasks and invisible work (Wang, Waldman, and Ashforth 2019).

Our participants often spoke about reputation and accountability with respect to AI-augmented writing tasks. For example, Participant U thinks that AI writing looks impressive, but it is not usable without considerable effort.

> You get something generated, it looks great, and you might say, this is so good, it's so helpful. But when you actually have to start using it, when you actually have to send it to your manager, once you start trying to do something with it, you realize, "oh, this looked great, it was a great structure, but I have to change so much to get it to what I need that it's actually not that helpful." But it looked great!

They spend a lot of time revising and editing their work to improve it beyond the AI baseline before they send it to their peers and managers. Participant N also talked about spending a lot of time "editing bad content" that AI had generated.

Other participants also commented on being accountable for the citations that research agents were finding. Participant R spoke about not being able to assess the quality of papers that were outside their fields, which may affect how colleagues in that field assess the value of their work:

> I find myself more asking the [AI research tools] to find papers relevant to an area I'm not an expert in. But then I can't really judge whether it found the relevant papers.

Several participants said they were willing to take additional time to learn about a topic and revise AI-generated content accordingly in order to hold themselves accountable for the work they are producing, including citing research outside their field. These results are in line with work showing that people still feel able to take control of the writing process and do not feel as influenced by AI's suggestions (Bhat et al. 2026; Park et al. 2026).

However, this finding was not universal. Research has shown that people who use AI to help in their decision-making feel less responsible for their decisions or results (Mendel et al. 2025). Some participants felt people were more tolerant of AI-written content and that they, as writers, were not subjected to the same amount of scrutiny for their writing as they used to be. Participant L noted that they would be less likely to use AI for research tasks if they knew they would be held accountable for the content:

> If I expected that there might be additional questions about a certain citation and an expectation for me to be the expert, I wouldn't be as comfortable using [AI research tools].

Fundamentally, they felt that their reputations were no longer on the line as they write more documents that fewer people read. Readers have the power to reduce the amount of low-quality AI content and increase productivity by holding writers accountable and AI content to high expectations (Niederhoffer et al. 2025).

## 5. Personal Growth and Learning

*are required to acquire new AI skills, and AI is also helping workers master new capabilities as informal learning from peers may be harder to come by.*

Using AI as a personalized learning tool is not new, AI-driven tutoring systems have been explored for decades (Nwana 1990). However, the advancement in chat interfaces and deep models has renewed the interest in such tools in schools (Alam 2023) as well as at work (Poquet and De Laat 2021). One recent literature review found that nearly half of the papers published on AI in the workplace focus on learning and training (Pereira et al. 2023).

Informal learning is proving to be critical as AI is constantly improving and the tools are changing. Spontaneous informal learning may be invisible to others but is critical to personal growth and career development (De Laat and Schreurs 2013). In the past, workers may have gotten these learning opportunities through interactions with their coworkers through collaborations or in meetings. Indeed, past research has shown that relationships are important for knowledge acquisition (Allen 1977; Granovetter 1973) and the following key relational characteristics impact information seeking and learning: (1) knowing what that person knows; (2) valuing what that person knows; (3) being able to gain timely access to that person's thinking; and (4) perceiving that seeking information from that person would not be too costly (Borgatti and Cross 2003).

However, with the advent of AI as an alternative information source, traditional mechanisms of learning and information seeking from peers appear to be impacted. Participant S talked about how they seek conceptual feedback from their peers and also want to provide feedback to others during meetings:

> I feel like for my team in particular, we still have a lot of meetings... you're not praised for asking questions in those meetings. It's more like "this is what we need to do."

As meetings are focused on the status of an increasing number of tasks that their team must complete, there is less space for teammates to talk about the details of the projects. While they could follow up after the meeting, asking the question during the meeting allows the entire team to learn at once and benefit from others' questions.

While we saw many examples of participants using AI to augment their own skills (e.g., designers vibe-coding), we also saw some participants opportunistically learning new skills from AI interactions. Participant D called it "a kind of personalized learning experience" that they find "really, really useful." Participant O talked about learning from AI to broaden their skill set. Rather than merely using the output of code writing, for example, they ask questions and probe the AI to understand what it is doing. They discussed several new technologies that they had picked up in recent months, and took pride in their ability to help their peers with a wider range of tasks.

> And it's not that I let it create and then I just go blindly trust it, right? Once it creates, I'm trying to understand and if I don't, I ask it what is it doing and it tells me what it's doing. So that learning also it is helping me.

Overall, participants who used AI for learning were satisfied with their interactions and felt empowered by their new skills despite growing concerns that AI inhibits on-the-job learning due to deskilling, pessimism about the future of work, and burnout (Li et al. 2023).

Finally, many participants discussed opportunities to learn about AI and to learn different tool chains for completing their work. While some said that learning sessions were very helpful, others also expressed frustration at the speed of the tools changing. Participant N said "It just feels like I could go learn something tomorrow and it'd be stale till the next day." Identifying critical skills that will continue to be of use throughout the shift to AI can help workers feel more willing to learn and less exasperated by the churn.

## 6. Instructional Support and Feedback

*from peers is dwindling and leading to a loss of social support, though AI's ability to provide help for many tasks creates a sense of independence.*

Task and instructional support, such as providing help or feedback, are critical aspects of social support at work; they are the strongest predictors of both job satisfaction and job tenure (Harris, Winskowski, and Engdahl 2007). Research shows that support networks beyond the traditional hierarchical organizational charts at work are responsible for the majority of work getting done (Cross, Borgatti, and Parker 2002; Cross and Parker 2004). Additionally, informal networks are critical for employee engagement, career development, and innovation at organizations (Farmer 2017). Giving and receiving help from other people is slow and takes time, and yet substituting AI for human connection may result in both an individual's feeling of work loneliness (Hadley and Wright 2024, 2026) and also the erosion of organizational support systems (Frögéli, Jenner, and Gustavsson 2023; Ehsan et al. 2026).

Our participants expressed a sense of self-sufficiency using AI tools to support their work. Participant F spoke about using AI for writing and finding work information:

> I'm probably a lot more independent because I reach out to AI for help as opposed to like a manager or like a team member or something like that.

Similarly, Participant L also appreciated AI's speed and ability to provide feedback rather than depending on a colleague's "eyes" for revision:

> [Previously] I was more prone to giving [my writing] to somebody else, either like a boss or a trusted peer to read through it. I find that I cut that out. [With AI] I will say especially in a quick turnaround, it's really nice to feel like you've had another set of eyes, so to speak, on it even though they're definitely not eyes.

Several participants spoke about the bottleneck of waiting for a peer during writing sessions, and that AI tools reduced the writing iteration time which allows them to produce more artifacts more quickly.

While AI tools are helping people feel more independent, we also saw indications that our participants were missing important professional networking and connection opportunities with their colleagues. Past research has shown that relationships with peers are critical for obtaining information (Krackhardt and Hanson 1993), learning to perform one's job (Mitschelen and Kauffeld 2025), and collectively solving complex tasks (Kahn 1990). Several participants reported feeling frustrated as they attempted to reach out to a colleague for help, only to be referred directly to AI. This is particularly important as participants join a new team and need to form new support networks, but also extends beyond onboarding.

Participant Q specifically mentioned social isolation with respect to the lack of collegial support:

> I definitely have felt more socially isolated here. . . . I will say here, typically the response that I get from someone is they'll plug my question into this chat and say, here's what this chat told me... So it feels inappropriate to really take up someone's time that way, because, it's like "let me Google that for you."

Similarly, Participant R reflected on the use of AI chats as the source of institutional knowledge and the lack of help or support if the chat does not have the answer.

> You're kind of expected to like ask [AI chat] first, and then sometimes even when you ask someone else, they ask [chat] and nobody really knows the answer who to talk to anymore. And if [chat] is not helpful, well, you just don't have any help how to do it.

These interactions may indicate that more senior co-workers feel that they do not have time to help their junior colleagues due to the increased intensity of AI work, or it may indicate that they have lost some of the institutional knowledge as AI has been substituted for more information-finding tasks at work. Either way, AI is having an impact on workers' abilities to find help to complete their tasks efficiently and we expect to see more downstream effects of the professional network breakdown in the future.

### 7. Task Support and Collaboration

*are tougher when AI makes work easier to do alone, especially since workers are finding that AI agents can be treated like peers or even supervisors at times.*

Task support extends beyond feedback. Workers form strategic partnerships and collaborations to improve their work output (Cross, Borgatti, and Parker 2002). There was a general sense among our participants that the speed of work is changing and is affecting their and their ability to work together deeply. Participant S talked about how the faster pace of work makes it hard to collaborate:

> It feels like a lot of our tasks are now more isolated than they used to be because it is easier for probably one person to do something than to collaborate on it ... And I do think it makes collaboration harder when iteration becomes faster too. Because you can start to diverge easily.

Although it is more challenging to collaborate with human peers, our participants described the interactions and teammate relationships they are building with AI. For example, Participant U explained that some of their team artifacts used to take weeks to produce and required iteration and consensus with a colleague that may not be supportive:

> Before [AI] I didn't have a thought partner. But it was a two-week, like, crazy process where you would partner with someone who would be like a sounding board. You would do iterations and it was really, really hard to get something finalized just because you were always back and forth with a person. With AI, it is a lot easier because it is just you and the AI and you can sort of zero in. Now the bad part is that the AI does not, usually the AI agrees with you.

Interestingly, they feel that AI is more of a partner than their human teammate, implying that their teammate's push-back was detrimental to the writing progress. But at the same time, they acknowledge that the AI is more sycophantic, and now they do not have the benefit of a devil's advocate to question the argument and iterate to make it stronger. Sycophancy in AI has been shown to be harmful to judgments and decision-making (Cheng et al. 2026).

Participant D reflected on their experience trying to derive a mathematical formulation with an AI agent, as they argued back and forth about whether different terms in the equations canceled each other out:

> It helped me, like, check some of the math... And it was really more like discussing on a board with my supervisor or colleague.

Participant P also uses AI agents as collaborators, though they turned to AI not just for the capabilities but because they found that they did not have as much peer support:

> I didn't really have anyone to talk to, like when I want to brainstorm something . . . But then now I can just, you know, back and forth [with AI], like brainstorming ideas, forming these ideas, explore possibilities out there. So yeah, it's really just amazing.

In these cases, the participants described their AI use as going beyond a typical assistant relationship and towards a collegial one where the AI agent has deep knowledge about the topic and is capable of having a peer discussion. Especially when knowledgeable colleagues are spread thin throughout an organization, having the ability to use AI to confirm assumptions, check arguments, and act as a contributor to novel work is key to producing high quality work and helping them grow. Both participants were happy with their interactions with the AI in these ways, despite the risk of losing out on professional collaborations.

When talking about their job role and job security, Participant P also spoke about the differences between their more senior colleagues and the junior ones: "I think the senior folks, they do have more expertise and also contacts in the area which is not replaceable by AI." They acknowledged that their lack of connections and junior status may make their job security more tenuous. One approach to helping build professional networks is through mentoring (Higgins and Kram 2001). Mentorship relationships have been shown to be key for career growth in academic professions (Goerisch et al. 2019) as well as in industry. Mentors can help advocate for their mentees, build their networks, and find opportunities to help them demonstrate their skills so they can grow their careers.

## Discussion and Conclusion

Our participants' invisible work patterns have been significantly impacted by AI, including their communication, collaboration, role clarity, task and instructional support, and information sharing. Broadly, we found that workers' interactions with their peers have shifted due to reliance on AI for help and information support and their ability to produce AI-augmented deliverables. With the intensity and breadth of work increasing, workers also feel more empowered to work independently using AI without having to manage the friction of sharing and collaborating with others. These changes will have widespread impact on the organizational culture and support systems that are necessary to help workers come together to build products effectively and efficiently while growing their professional careers and networks. We propose opportunities for AI companies, organization leaders, and individuals to build additional mechanisms to support invisible work.

**AI is not just changing work, it is changing how we interact with and support each other in key ways.** For some professionals like our designers, who had been struggling to find common ground with software developers who needed to implement their interaction designs, AI has given them a new tool to communicate their intentions. For other workers, who used to share their nuanced findings through presentations and long documents, the reduced attention span of their peers seems to be limiting their opportunities for clearly communicating their work. While some responsibility rests on the writers, several participants noted that it is extremely challenging for readers to maintain situational awareness while consuming such a small portion of the work. Individuals should be aware of the potential costs, including misunderstandings and slowed productivity, of not reading, providing feedback, and accurately interpreting the artifacts that their peers are spending time producing.

Our findings also show a marked lack of help and feedback provided by peers. It can feel uncomfortable and can take time to ask for help from a colleague, rather than turning to AI for faster, less-judgmental feedback. But we also had participants who do seek help but cannot find someone available, willing, and capable of provide the necessary support. Managers must be responsible for creating the psychological safety among their teams to be able to ask for help, and also to encourage their senior colleagues to prioritize providing help to others. Additionally, AI companies could take steps to name the authors of content that is referenced or linked to AI output rather than only linking the location, helping to draw the social connection between creator and user of the content.

**AI is empowering people to be independent at work and, at the same time, is dismantling the existing mechanisms that are used to drive productivity in large organizations.** Our participants talked about the ability to create more and wider varieties of contributions using AI. However, it can still be challenging for AI to leverage or contribute to the work of others, so many of the AI-augmented contributions are created independent of institutional knowledge from colleagues, such as vibe-coded demos compared to legacy code bases. Contributing to and leveraging existing work products reduces the likelihood of duplicated efforts which can be both expensive and waste time. Participants also discussed the shift in collaboration opportunities. Colleagues build rapport and make professional connections that can be utilized later to drive clarity on product features and find out information from other groups. When junior co-workers collaborate with senior ones, they learn organizational knowledge as well as skills.

While the productivity seems faster because the creator does not have to fight with existing infrastructure or communicate clearly with collaborators, we believe that there will be eventual slow-downs as more time is spent integrating disparate pieces that were designed independently without each other in mind. Prior work found a similar AI-as-Amplifier paradox with respect to independence and deskilling (Ehsan et al. 2026).

**Notable increases in low-value AI content are, at least in part, driven by the eroding social structures that hold content creators responsible for their work.** Our participants all discussed their efforts to maintain accountability for the quality of their work despite the organizational pressures to work more quickly and on a broader set of tasks. However, they also overwhelmingly felt that the quality bar was lower for AI-augmented content than it had been prior to the introduction of AI in the workplace, and therefore did not feel as compelled to revise and check their work. At the same time, our participants noted that many people were using AI to summarize content that they generated, potentially missing key points that the author was trying to convey in the longer context. This has the potential to become a vicious cycle in which AI is generating content and then summarizing it again, rather than merely generating the summary from the beginning. We believe that to reduce the occurrence of low-value content, readers must hold their colleagues accountable for the work they produce, maintaining a level of quality by asking questions and providing feedback.

Finally, **AI is impacting different workers differently.** Our participants experienced AI transformation differently based on their experience level, their discipline, and even their personal preferences on their individual work tasking. Participants who were eager to expand their task scope in the ways that AI allowed felt more empowered than those participants who felt encumbered by AI technology that could not meet their needs or their high expectations. Junior participants who had joined the company more recently were more impacted by the lack of social support because they could not rely on their professional connections. And the more computationally-focused disciplines reported less direct productivity impact of AI tools compared to the designers and user researchers, although their invisible work like collaboration were notably affected. **This inequality is an organizational risk and leaders need to be aware that AI is creating conditions that hinder professional and career growth** and develop programs to support all workers through their AI transformation.

Moving forward, AI companies must focus efforts on identifying ways to increase social support and collaboration through their AI tools rather than eliminating it. Organization leaders, managers, and individuals must also identify AI-driven changes in their own behavior and those of their peers, and implement programs to support invisible work that helps make their organizations run smoothly.

## Ethical Considerations

This study was approved by our institution's research ethics committee. Participants were recruited through company email lists and participation was voluntary. Before starting the interview, participants provided their consent after being informed of the purpose of the study, task requirements, expected time commitment, and their right to skip questions or withdraw at any time without penalty. The study relies on open ended responses from the participants. Although we had a fixed set of questions for all participants, occasionally additional questions were asked based on the responses. The interviews took no more than an hour and the participants were not compensated. Interview responses were transcribed live but not recorded, and relevant portions of those interviews were saved in an anonymized format for further analysis. No personally identifying information was retained. The findings from the study are based on interview responses from a small sample for four job functions and broader generalizability should be treated with caution without larger scale confirmation through further interviews or surveys.